\pdfoutput=1

\documentclass[10pt,twocolumn,letterpaper]{article}

\usepackage[pagenumbers]{cvpr} 

\usepackage{graphicx}
\usepackage{amsmath}
\usepackage{amssymb}
\usepackage{booktabs}
\usepackage{color}
\usepackage{verbatim}
\usepackage[pagebackref,breaklinks,colorlinks]{hyperref}
\usepackage{multirow}
\usepackage{marvosym}

\usepackage[capitalize]{cleveref}
\crefname{section}{Sec.}{Secs.}
\Crefname{section}{Section}{Sections}
\Crefname{table}{Table}{Tables}
\crefname{table}{Tab.}{Tabs.}

\def\cvprPaperID{2513} 
\def\confName{CVPR}
\def\confYear{2023}

\begin{document}

\title{Revisiting Temporal Modeling for CLIP-based Image-to-Video \\ Knowledge Transferring}

\author{{Ruyang Liu\footnotemark[1]}~$^{1 \Diamond}$
    ~~~{Jingjia Huang\footnotemark[1]}~$^{2}$
    ~~~{Ge Li \footnotesize{\Letter}}$^{1}$
    ~~~Jiashi Feng$^{2}$ 
    ~~~Xinglone Wu$^{2}$
    ~~~Thomas H. Li$^{1}$ \\
    {\small $^1$School of Electronic and Computer Engineering, Peking University
    ~~~$^2$ByteDance Inc~~~}\\
{\tt\small \{ruyang@stu,geli@ece,thomas@\}.pku.edu.cn~~~\{huangjingjia,jshfeng,wuxinglong\}@bytedance.com}\\  
}
\maketitle

\begin{abstract}
   Image-text pretrained models, e.g., CLIP, have shown impressive general multi-modal knowledge learned from large-scale image-text data pairs, thus attracting increasing attention for their potential to improve visual representation learning in the video domain. In this paper, based on the CLIP model, we revisit temporal modeling in the context of image-to-video knowledge transferring, which is the key point for extending image-text pretrained models to the video domain. We find that current temporal modeling mechanisms are tailored to either high-level semantic-dominant tasks (\textit{e.g.}, retrieval) or low-level visual pattern-dominant tasks (\textit{e.g.}, recognition), and fail to work on the two cases simultaneously. The key difficulty lies in modeling temporal dependency while taking advantage of both high-level and low-level knowledge in CLIP model. To tackle this problem, we  present Spatial-Temporal Auxiliary Network (STAN) -- a simple and effective temporal modeling mechanism extending CLIP model to diverse video tasks. Specifically, to realize both low-level and high-level knowledge transferring, STAN adopts a branch structure with decomposed spatial-temporal modules that enable multi-level CLIP features to be spatial-temporally contextualized. We evaluate our method on two representative video tasks: Video-Text Retrieval and Video Recognition. Extensive experiments demonstrate the superiority of our model over the state-of-the-art methods on various datasets, including MSR-VTT, DiDeMo, LSMDC, MSVD, Kinetics-400, and Something-Something-V2. Codes will be available at \href{https://github.com/farewellthree/STAN}{https://github.com/farewellthree/STAN}
\end{abstract}

\section{Introduction} \label{sec:intro}
    Recent years have witnessed the great success of image-text pretrained models such as \textit{CLIP} \cite{radford2021learning}. Pretrained on over 400M image-text data pairs, these models learned transferable rich knowledge
    for various image understanding tasks. Similarly, video domains also call for a \textit{CLIP}-like  model to solve downstream video tasks.
    However, it is hard to get a pretrained model as powerful as \textit{CLIP} in the video domain due to the unaffordable demands on computation resources and the difficulty of collecting video-text data pairs as large and diverse as image-text data. Instead of directly pursuing video-text pretrained models \cite{huang2022clover, miech2019howto100m}, a potential alternative solution that benefits video downstream tasks is to transfer the knowledge in image-text pretrained models to the video domain, which has attracted increasing attention in recent years \cite{wang2021actionclip,ni2022expanding,pan2022st,luo2021clip4clip,fang2021clip2video,gao2021clip2tv}. 
    
    Extending pretrained 2D image models to the video domain is a widely-studied topic in deep learning \cite{carreira2017quo,bertasius2021space}, and the key point lies in empowering 2D models with the capability of modeling temporal dependency between video frames while taking advantages of  knowledge in the pretrained models. 
    In this paper, based on \textit{CLIP} \cite{radford2021learning}, we revisit temporal modeling in the context of image-to-video knowledge transferring, and
    present Spatial-Temporal Auxiliary Network (STAN) -- a new temporal modeling method that is easy and effective for extending image-text pretrained model to diverse downstream video tasks.

    \begin{figure*}[t] 
    \setlength{\abovecaptionskip}{-0.05 cm}
    \setlength{\belowcaptionskip}{-0.3cm}
    \centering
    \includegraphics[width=1\textwidth]{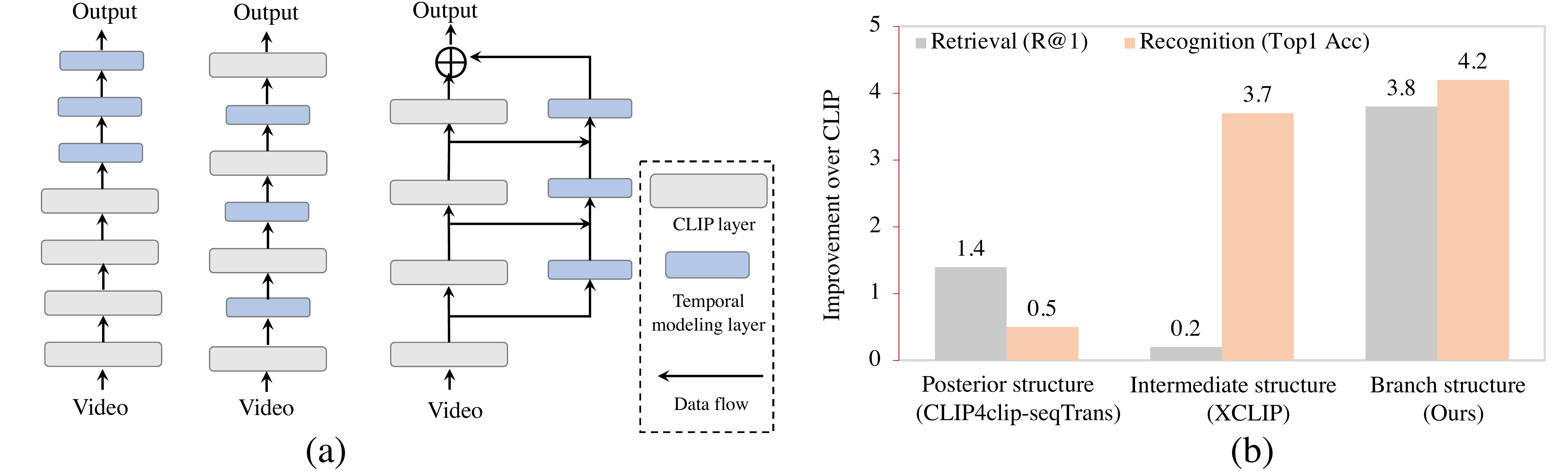} 
    \caption{(a) Illustration of temporal modeling with posterior structure (left), intermediate structure (middle) and our branch structure(right). (b) Performance comparison among the posterior structure based CLIP4clip-seqTrans \cite{luo2021clip4clip} , intermediate structure based XCLIP \cite{ni2022expanding} and our branch structure based STAN. We take the \textit{CLIP} model with a naive mean pooling to aggregate the features of all frames into video representations as the baseline. We present the improvement brought   by different methods over this  baseline w.r.t.\  Recall@1 on MSRVTT for video-text retrieval and Top-1 accuracy on Kinetics-400 for video recognition.}
    \label{img_intro}
  \end{figure*}
  
    
    We find that current efforts on empowering \textit{CLIP} with temporal modeling capability can be roughly divided into posterior structure based methods and intermediate structure based methods as shown in Fig.~\ref{img_intro}(a). Posterior structure based methods \cite{luo2021clip4clip, gao2021clip2tv, fang2021clip2video} employ a late modeling strategy, which take \textit{CLIP} as a feature extractor and  conduct temporal modeling upon the embeddings of video frames extracted independently from \textit{CLIP}. Upon the highly semantic embeddings, though the structure is beneficial for transferring the well-aligned visual-language representation (\emph{i.e.,} high-level knowledge) to downstream tasks, it hardly captures the spatial-temporal visual patterns (\emph{i.e.,} low-level knowledge) among different frames, which is important for video understanding. As shown in Fig.~\ref{img_intro}(b), compared to the \textit{CLIP} baseline that employs a naive mean pooling to aggregate the features of all frames to obtain a video representation, the performance improvement brought   by the typical posterior structure, \emph{i.e.} CLIP4clip-seqTrans \cite{luo2021clip4clip} is trivial, especially on the video action recognition task where spatial-temporal visual patterns are crucial. 
    
    In contrast to posterior structure based methods, intermediate structure based methods \cite{ni2022expanding, pan2022st, bertasius2021space}  strengthen the spatial-temporal modeling capability of \textit{CLIP} via plugging temporal modeling modules directly between \textit{CLIP} layers, and achieve 3.7\% improvement over the baseline on the video action recognition task. However, we find that inserting additional modules into \textit{CLIP} would impact the pretrained high-level semantic knowledge in the model, which only outperforms the baseline by 0.2\% on   the video-text retrieval tasks. Therefore,    modeling temporal dependency while taking advantage of knowledge in different levels of representation is important for extending the \textit{CLIP} model to the video domain.

  Unlike the above methods, inspired by FPN \cite{Lin_2017_CVPR} that introduces a branch network to strengthen multi-level representation learning for CNNs, our proposed STAN employs a new branch structure  outside of the visual backbone,  as shown in Fig.~\ref{img_intro}(a).
  Thanks to the branch structure, STAN augments the features of video frames with spatial-temporal contexts at different \textit{CLIP} output levels  without affecting the forward-propagating of \textit{CLIP} itself. Thus, it is able to take advantage of both high-level and low-level knowledge in the pretrained model simultaneously, and effectively extends \textit{CLIP}  to diverse downstream video tasks.    STAN  consists of multiple layers with a spatial-temporal separated design. Specifically, the layer operates spatial-temporal modeling via alternatively stacking two separate modules – an intra-frame module and a cross-frame module, which enables the layer to boost the performance of model via reusing the pretrained parameter of \textit{CLIP} layers to initialize the intra-frame spatial modules. We further investigate two instantiations of cross-frame modules, \emph{i.e.,} the self-attention-based module and 3D convolution based module, to facilitate the comprehensive understanding of STAN in different implementations.
    We evaluate our STAN on both the high-level semantic-dominant task (\textit{i.e.,} video-text retrieval) and low-level visual pattern-dominant task (\textit{i.e.,}, video recognition), trialing our methods from the two different perspectives. Extensive experiments demonstrate our expanded models are generally effective on the two different tasks. For video-text retrieval, we surpass the CLIP4clip by +3.7\%, +3.1\%, and +2.1\% R@1 on MSRVTT, DiDemo, and LSMDC. For video recognition, we achieve competitive performance on Kinetics-400, with $88\times$ fewer FLOPs than Swin3D-L \cite{liu2021video} and improve \textit{CLIP} baseline by 20\%+ on Something-Something-V2.

    Our main contributions are summarized as: (1) we revisit temporal modeling in the context of image-to-video knowledge transferring and figure out that the key challenge lies in modeling temporal dependency while taking advantage of both high-level and low-level knowledge; (2) we propose Spatial-Temporal Auxiliary Network (STAN) -- a new branch structure for temporal modeling, which facilitates   representation learning of video frames with including spatial-temporal contexts at different levels and better transfer the pretrained knowledge in \textit{CLIP} to diverse video tasks; (3) our method achieves competitive results on both video-text retrieval and video recognition tasks compared to SOTA methods.
    
\section{Related Work} \label{sec:RWork}    
    \noindent \textbf{Visual-Language Pre-Training.}
    Visual-Language pretraining has drawn growing attention in past years \cite{miech2019howto100m, sun2019learning, sun2019videobert}. Recently, the contrastive language-image pretraining on web-scale data \cite{radford2021learning, yuan2021florence, jia2021scaling} achieves great success for its remarkable performance when transferring to various downstream tasks. One of the most famous works is the \textit{CLIP}  \cite{radford2021learning}, which has revealed surprising capacities of zero-shot recognition and domain generalization  \cite{zhou2022learning, patashnik2021styleclip, luo2021clip4clip}. However, language-video datasets suffer from either finite scale \cite{bain2021frozen} or noisy subtitle annotations \cite{xue2022advancing, miech2019howto100m} as well as expensive computation consumes, hence the limited improvement from the pretraining. Thereby, efforts are made \cite{luo2021clip4clip,gao2021clip2tv,fang2021clip2video,min2022hunyuan_tvr,wang2022disentangled,ni2022expanding,pan2022st, liu2022contextual} to adapt the language-image pretraining models to video tasks, which even get better results than methods pretrained on video datasets.

    \vspace{0.3em}

    \noindent \textbf{\textit{CLIP} for Video-Text Retrieval.} 
    \textit{CLIP} contains rich vision-text aligned knowledge, which is favoured by the video-text retrieval task. Early works \cite{luo2021clip4clip, gorti2022x, zhao2022centerclip, gao2021clip2tv,fang2021clip2video} try to add temporal modeling modules as a posterior structure to \textit{CLIP}, \textit{e.g.}, the sequential transformer in \cite{luo2021clip4clip} and the temporal difference transformer in \cite{fang2021clip2video}. Despite the progress they have made, the temporal modeling is limited in high-level embeddings and not effective enough as shown in Fig. \ref{img_intro}(b). There are also some works that modify \textit{CLIP} from the perspective of disentangling and multi-level representation interaction \cite{gorti2022x, wang2022disentangled, min2022hunyuan_tvr}, and achieve general advancement on various video-text retrieval datasets. However, these methods can only be applied to tasks with sentence input (\textit{i.e.}, multimodal tasks), and are not suitable for recognition tasks. In contrast, our method advances the retrieval as well as other video tasks through effective temporal modeling.

     \vspace{0.3em}

    \noindent \textbf{\textit{CLIP} for Video Recognition.} 
    Compared to the retrieval task, the recognition task requires a model to better modeling the dynamic visual patterns in videos, where the visual patterns in \textit{CLIP} learnt from large-scale image-text pretraining data are valuable. Therefore, there are numbers of works migrating the \textit{CLIP} to video recognition \cite{bain2021frozen, ni2022expanding, pan2022st, wang2021actionclip, ju2021prompting}. Some of them focus on the prompting or sampling modeling \cite{bain2021frozen, ju2021prompting, ni2022expanding}, and others \cite{ni2022expanding, pan2022st, wang2021actionclip, bertasius2021space} design temporal modules as a intermediate structure illustrated in Fig \ref{img_intro}(a). Ni \textit{et al} \cite{ni2022expanding} insert the message token to input frame tokens to capture sequence information. Pan \textit{et al} \cite{pan2022st} develop 3D convolution modules as adapters plugged between the \textit{CLIP} layers. Unlike the aforementioned methods, we propose a branch structure based method for better transferring image-text model to the video domain.

\section{Methodology} \label{sec:method} 
\subsection{Motivation and Overview}

     \textit{CLIP} is a large-scale image-text pretrained model which learns general multi-modal knowledge from 400 million image-text pairs. It consists of two encoders for the extraction of image and text representation respectively, where the visual encoder is composed of a stack of transformer-based \cite{vaswani2017attention} encoder layers. From the bottom to the top of layers, the visual encoder gradually learns the visual patterns at different levels of abstraction \cite{yuan2021tokens}, and at last outputs high-level visual embedding semantically aligned with the corresponding embedding in the text modality.
    
    \textit{CLIP}-based image-to-video transferring aims to improve the learning of video representation with the pretrained knowledge in \textit{CLIP}, where the key point lies in empowering the image encoder in \textit{CLIP} with the capability of modeling temporal dependency between video frames. Current works typically introduce extra modules as a posterior or intermediate structure of \textit{CLIP} visual encoder for explicitly temporal modeling towards different downstream video tasks. For high-level semantic knowledge dominant tasks, \emph{e.g.,} video-text retrieval,  the posterior structure takes advantage of the pretrained visual-language alignment knowledge via operating temporal modeling upon the outputs of \textit{CLIP}. As for visual pattern dominant tasks, \emph{e.g.,} video action recognition, the intermediate structure benefits from the pretrained visual patterns knowledge in \textit{CLIP}, named as low-level knowledge, and empowers the encoder with the capability of learning spatial-temporal patterns from the video. Nevertheless, the posterior structure and the intermediate structure based temporal modeling methods fail to transfer the high-level and low-level knowledge to the video domain simultaneously.
    
\begin{figure*}[t] 
    \setlength{\belowcaptionskip}{-0.3cm}
    \centering
    \includegraphics[width=1\textwidth]{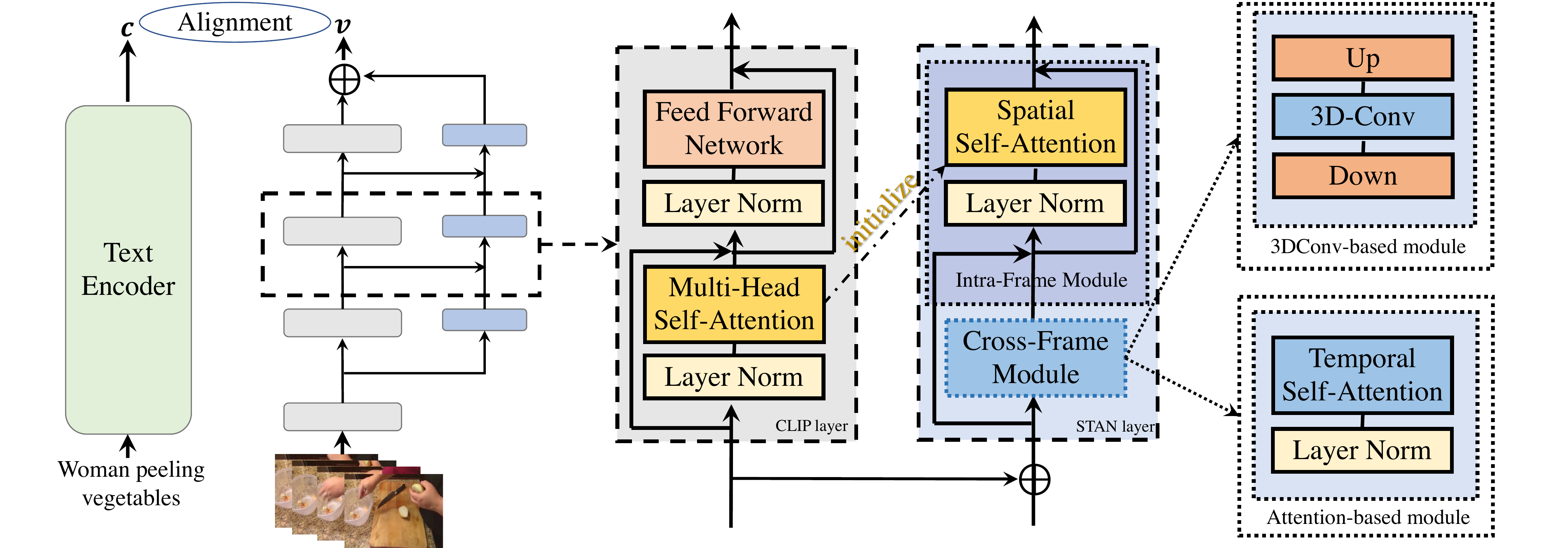} 
    \caption{The overview of our proposed STAN architecture, including the global overview of our backbone (left), details of the internal structure of our spatial-temporal module (middle), and implementations of the cross-frame module (right).}
    \label{img_method}
  \end{figure*}

    Therefore, we propose Spatial-Temporal Auxiliary Network (STAN), a new temporal modeling mechanism for \textit{CLIP}-based image-to-video knowledge transferring. As shown in Fig. \ref{img_method}, STAN consists of a stack of $K$ spatial-temporal layers and acts as a branch structure beside the \textit{CLIP} visual encoder.
    Given a video with $T$ frames, the frames are fed into the \textit{CLIP} visual backbone to obtain intermediate outputs at last $K+1$ level of \textit{CLIP} layers. We denote the outputs of the k$th$ selected \textit{CLIP} layer as:
    \begin{equation}
        V^k=\{f_{i,j}^{k} \in \mathcal{R}^{D}|i\in [1,T], j \in [0,L]\},
    \end{equation}
     which is a visual embedding sequence of the video where $T$, $L$ and $D$ are the frame number, per-frame patch number and embedding dimension, respectively. In $V^k$,  $f_{i,0}^{k}$ indicates the embedding of [CLS] token in the i$th$ frame of the video while $f_{i,j>0}^{k}$ represents the visual embedding of j$th$ patch in the frame.
    Then, we feed each intermediate output $V^k$ into the corresponding level of layer in STAN for the modeling of spatial-temporal correspondence between video frames. At last,  we fuse the  output of the last \textit{CLIP} layer  with the output of STAN to get the final representation of the video.

    Compared to the posterior structure based methods, STAN operates spatial-temporal modeling on multi-level \textit{CLIP} representations and thereby is able to better capture the visual dynamics information in the video. Meanwhile, unlike previous intermediate structure based methods, which insert additional modules into \textit{CLIP} visual encoder, the branch structure of STAN avoids destroying the inherent structure of the visual encoder and thereby protect the pretrained knowledge, especially the high-level visual-text alignment knowledge in \textit{CLIP}.
  
\subsection{Spatial-Temporal Auxiliary Network}
    
    STAN consists of a stack of $K$ spatial-temporal layers, where the input for each layer is constructed based on the output of a \textit{CLIP} visual layer. For the k$th$ layer in STAN, its input is an embedding sequence of the video denoted as:
    \begin{equation}
        V'^k=\{f'^{k}_{0,0},f'^{k}_{1,1},..,f'^{k}_{1,L},..,f'^{k}_{T,1},..,f'^{k}_{T,L}\},
    \end{equation}
     where $f'^{k}_{0,0}$ is the embedding of [CLS] token for the whole video while others are embedding of image patches in different frames. The output of the STAN layer is also an embedding sequence with the same size as its input, which is denoted as:
     \begin{equation}
    \widetilde{V}^k = \{\widetilde{f}^{k}_{0,0},\widetilde{f}^{k}_{1,1},..,\widetilde{f}^{k}_{1,L},..,\widetilde{f}^{k}_{T,1},..,\widetilde{f}^{k}_{T,L}\},
     \end{equation}
    
    At the first STAN layer, to construct its input from $V^1$, we first average the embedding of [CLS] tokens in each frame as a new embedding $f'^{1}_{0,0}=\frac{1}{T}\sum_{i \in T}f^{1}_{i,0}$, and then update patch embeddings in $V^1$ with spatial and temporal position embeddings as: 
    \begin{equation} 
             f'^1_{i,j} = \mathrm{Dropout}(f_{i,j}^{1} + \mathrm{Pos_t}(t)+\mathrm{Pos_s}(j)),
    \end{equation}
    where $j>0$ while $\mathrm{Pos_t}$ and $\mathrm{Pos_s}$ are the learnable embeddings for the temporal and spatial position of each patch. 
    For the other layers in STAN, the input $V'^k$ is constructed based on the output from the previous STAN layer $\widetilde{V}^{k-1}$ and \textit{CLIP} output $V^k$ as follows:
    \begin{gather} 
    f'^{k}_{0,0}  = \widetilde{f}^{k-1}_{0,0} + \mathrm{W}^{k}_{proj} \frac{1}{T}\sum_{i \in T}f^{k}_{i,0}  , \\
    f'^{k}_{i,j}  = \widetilde{f}^{k-1}_{i,j} + \mathrm{W}^{k}_{proj} f^{k}_{i,j}  , 
    \end{gather}
    where $i\in[1,T], j\in[1,L]$, and $\mathrm{W}^{k}_{proj} \in \mathbf{R}^{D \times D}$ is a projection layer.

    Given the input embedding sequence of the video, STAN layer learns the spatial-temporal information among the video frames. As shown in Fig. \ref{img_method}, it operates temporal modeling via alternatively stacking two separated modules -- an intra-frame module  and a cross-frame module. Thanks to the separated design, we are able to reuse the structure in \textit{CLIP} visual encoder layer as our intra-frame spatial module and initialize it with the pretrained model, which effectively improves the performance on downstream tasks.
    Same as \textit{CLIP}, the intra-frame module is also a self-attention block responsible for spatial modeling. For simplicity, we omit the superscript of embedding and denote the embeddings in frame $i$ as $X_i \in \mathbf{R}^{(L+1)\times D}$, where the embedding of [CLS] token in the video is duplicated and concatenated with patch embeddings. In each frame, the spatial module updates embeddings via self-attention:
    \begin{small}
    \begin{equation}
        \hat{X}_i = \mathrm{softmax}(X_i\mathrm{W_Q} (X_i\mathrm{W_K})^\mathrm{T} / \sqrt{D})(X_i\mathrm{W_V}) + X_i,
    \end{equation} 
    \end{small}
    where $\mathrm{W_Q} / \mathrm{W_K} / \mathrm{W_V}$ indicate the linear projections for the query, key and value in self-attention layer of the spatial module. After that, the duplicated [CLS] embeddings in each frame are averaged as the video [CLS] embeddings.
  
    The cross-frame module is responsible for temporal modeling. For simplicity, we omit the superscript of embedding and denote the collection of j$th$ patch embeddings in different frames as $Y_j \in \mathbf{R}^{T \times D}$. At each spatial position, the patch embeddings are updated as $\hat{Y}_j = Temp(Y_j)$, where $Temp()$ indicates the message passing strategy across temporal dimension which can be instantiated in different ways. In the next section, we present a self-attention-based cross-frame module and a 3D convolution-based cross-frame module, and study the performance of the two instantiations.


\subsection{Temporal Modeling in STAN} \label{tmSTAN}
     In deep learning, there are various ways to achieve temporal modeling, for example, 3D convolution \cite{carreira2017quo, tran2015learning}, temporal self-attention \cite{bertasius2021space, arnab2021vivit} and proxy tokens \cite{ni2022expanding, fan2021can}. In this paper, we  investigate two most popular instantiations of temporal modeling in the proposed framework, \emph{i.e.,} the self-attention based module and convolution based module, to facilitate the comprehensive understanding of STAN in different implementations.\\
    \noindent \textbf{Self-attention based module.} 
    Self-attention has a natural advantage in sequence modeling due to its global modeling capability. At each spatial position, the patch embeddings from  different frames are updated as: 
    \begin{gather} 
    \label{qkv}
    \hat{Y}_i = \mathrm{softmax}(Y_i\mathrm{W_Q} (Y_i\mathrm{W_K})^\mathrm{T} / \sqrt{D})(Y_i\mathrm{W_V}) + Y_i,
    \end{gather}
    where $\mathrm{W_Q} / \mathrm{W_K} / \mathrm{W_V}$ indicate the linear projections for the query, key, and value in self-attention layer of the cross-frame module. Through temporal attention, each patch is contextualized with temporal information at the same locations. 

    \vspace{0.2em}

    \noindent \textbf{Convolution based module.}
    Convolution operator has been widely adapted for effective temporal modeling in CNNs \cite{carreira2017quo, xie2018rethinking, tran2015learning}, e.g., C3D \cite{tran2015learning}, S3D \cite{xie2018rethinking}. Though self-attention gains increasing attention, convolution still owns the advantage of better local modeling and easier easier to converge. Therefore, we also implement the cross-frame module of STAN based on the convolution operator. Specifically, we stack the patch embeddings of the video to form a 3D feature cube $Y \in \mathbf{R}^{T\times W \times H \times D}$ and then update the features as follows:
    \begin{equation} 
             Y = Up(\mathrm{Gelu}(\mathrm{3DConv}(Down(Y)))) + Y,
    \end{equation}
    where the $Down()$ and $Up()$ are the point-wise convolution operators with channel size of $\frac{D}{8}$ and $D$, which reduce and restore the dimension of patch embeddings. As for the kernel size of 3D convolution, the dimensions for T, H, and W are set to 3, 1, and 1 respectively.

\section{Experiments} \label{sec:method} 
\subsection{Experiment Settings}
\noindent \textbf{Datasets.}
We evaluate our method on both the high-level semantic-dominant task \textit{i.e.,}, video-text retrieval, and low-level visual pattern-dominant task \textit{i.e.,}, video recognition, trialing our methods from the two different perspectives.
For video-text retrieval, we employ \textit{MSR-VTT} \cite{xu2016msr}, \textit{DiDemo} \cite{anne2017localizing} and \textit{LSMDC} \cite{rohrbach2017movie}; for video recognition, we adopt \textit{Kinetics-400} \cite{kay2017kinetics} and \textit{Something-Something-v2} \cite{goyal2017something}. 

\textit{MSR-VTT} is the most popular benchmark consisting of 10,000 YouTube videos with 20 captions for each video.
 \textit{DiDemo} contains 10,000 videos and 40,000 sentences with longer video duration than other retrieval datasets.
\textit{LSMDC} is a large-scale video-text retrieval benchmark with 118,081 videos from 202 movies, which is more diverse in concept and duration than other datasets. 

\textit{Kinetics-400 (K-400)} is a popular video action recognition dataset that has 260,000 videos with average 300 frames and 400 action classes.  \textit{Something-Something-v2 (SSv2)} is a video action recognition benchmark especially for temporal modeling, which contains 220,485 videos and 174 action classes. In \textit{K-400}, most of the action categories are biased to static scene context \cite{Sevilla-Lara_2021_WACV}.
In \textit{SSv2}, the classes of action are less relevant to the static scene context but closely related to the dynamic information in videos. 

\vspace{0.3em}

\noindent \textbf{Implementation Details.}
We set the number of STAN layers to 4 for all datasets except on SSv2 when it is set to 6. We employ the simple cross-entropy loss and NCE loss for fine-tuning on video recognition and video-text retrieval, respectively. Following previous work \cite{luo2021clip4clip}, we fine-tune the model with a frame number of 12 and a token length 32 for MSRVTT, LSMDC, K400, and SSv2. On Didemo where videos have a longer duration,  the frame number and token number are set to 64 and 64. The batch size is set to 128 for all datasets. We adopt Adam as our optimizer with weight decay of 0.02. The learning rates are initialized to 2e-6 and 2e-5 for parameters in \textit{CLIP} and parameters in STAN respectively, and then decay following a cosine annealing decay schedule. For more details and code, please refer to supplementary materials.

\subsection{Comparisons with State-of-the-art}\label{sec:sota}

\noindent \textbf{Video-Text Retrieval.} 
We compare our STAN with current SOTAs including both video-text pretrained and \textit{CLIP}-pretrained methods across different benchmarks. Comparisons on MSR-VTT, DiDemo abd LSMDC are reported in Table \ref{msrvtt}, \ref{didemo}, and \ref{lsmdc}, respectively. For \textit{CLIP}-pretrained methods, unless denoted with \emph{B/16}, all the methods are based on \textit{CLIP}-B/32. For our method, we report the results achieved by both \textit{CLIP}-B/32 and \textit{CLIP}-B/16, and denote STAN with self-attention and 3D convolution based inter-frame module as STAN-self and STAN-conv, respectively. 

\begin{table}[]
   \setlength{\abovecaptionskip}{0.cm}
   \begin{center}
     \caption{Comparisons on MSR-VTT \cite{xu2016msr}. We train on Training-9K and test on Test-1k-A. * means extra tricks (\textit{e.g.,} DSL \cite{cheng2021improving} and QB-Norm \cite{bogolin2022cross}) are utilized during inference. } 
     
     \footnotesize
     \label{msrvtt}
     \renewcommand\tabcolsep{9pt}
     \scalebox{0.92}{
     \begin{tabular}{ccccc} 
     \hline
     \multicolumn{1}{c}{Methods} & \multicolumn{1}{c}{R@1 $\uparrow$}  & \multicolumn{1}{c}{R@5 $\uparrow$} & \multicolumn{1}{c}{R@10 $\uparrow$} & \multicolumn{1}{c}{MdR $\downarrow$} \\ \hline 
     \multicolumn{5}{l}{\textit{Pretrained on large-scale video-text dataset}}  \\  
     \multicolumn{1}{l}{ClipBERT \cite{lei2021less}} & 22.0 & 46.8 & 59.9 & 6.0 \\
     \multicolumn{1}{l}{Frozen \cite{bain2021frozen}} &  31.0 & 59.5 & 70.5 & 3.0 \\
     \multicolumn{1}{l}{HD-VILA \cite{xue2022advancing}} & 35.6 & 65.3 & 78.0 & 3.0 \\
     \multicolumn{1}{l}{All-in-one \cite{wang2022all}} &  37.9 & 68.1 & 77.1 & - \\
    \multicolumn{1}{l}{BridgeFormer \cite{ge2022bridging}} &  37.6 & 64.8 & 75.1 & 3.0 \\
     \multicolumn{1}{l}{Clover \cite{huang2022clover}} &  38.6 & 67.4 & 76.4 & 2.0 \\ \hline 
     \multicolumn{5}{l}{\textit{CLIP pretrained}}  \\  
     \multicolumn{1}{l}{Clip4clip \cite{luo2021clip4clip}} & 44.5 & 71.4 & 81.6 & 2.0 \\
     \multicolumn{1}{l}{CenterCLIP \cite{zhao2022centerclip}} &  44.2 & 71.6 & 82.1 & 2.0 \\
     \multicolumn{1}{l}{CLIP2Video* \cite{fang2021clip2video}} &  47.2 & 73.0 & 83.0 & - \\
     \multicolumn{1}{l}{CAMoE* \cite{cheng2021improving}} &  47.3 & 74.2 & 84.5 & 3.0 \\
    \multicolumn{1}{l}{CLIP2TV-B/16 \cite{gao2021clip2tv}} &  49.3 & 74.7 & 83.6 & 2.0 \\
     \multicolumn{1}{l}{DRL-B/16* \cite{wang2022disentangled}} &  53.3 & \textbf{80.3} & 87.6 & 1.0 \\ \hline 
     \multicolumn{5}{l}{\textit{Our method}}  \\
     \multicolumn{1}{l}{STAN-self-B/32} &  46.9 & 72.8 & 82.8 & 2.0 \\
     \multicolumn{1}{l}{STAN-conv-B/32} &  46.6 & 72.8 & 82.2 & 2.0 \\
     \multicolumn{1}{l}{STAN-self-B/32*} &  49.0 & 74.8 & 83.5 & 2.0 \\
     \multicolumn{1}{l}{STAN-self-B/16} &  50.0 & 75.2 & 84.1 & 1.5 \\
     \multicolumn{1}{l}{STAN-self-B/16*} & \textbf{54.1} & 79.5 & \textbf{87.8} & 1.0 \\  \hline 
     
   \end{tabular}  }
   \end{center}
   \vspace{-1.2em}
\end{table}

\begin{table}[]
   \setlength{\abovecaptionskip}{0.cm}
   \begin{center}
     \caption{Comparisons on DiDemo \cite{anne2017localizing}. We concatenate all captions of a video into a single query. * means extra tricks (\textit{e.g.,} DSL \cite{cheng2021improving} and QB-Norm \cite{bogolin2022cross}) are utilized during inference.} 
     
     \footnotesize
     \label{didemo}
     \renewcommand\tabcolsep{9pt}
     \scalebox{0.92}{
     \begin{tabular}{ccccc} 
     \hline
     \multicolumn{1}{c}{Methods} & \multicolumn{1}{c}{R@1 $\uparrow$}  & \multicolumn{1}{c}{R@5 $\uparrow$} & \multicolumn{1}{c}{R@10 $\uparrow$} & \multicolumn{1}{c}{MdR $\downarrow$} \\ \hline 
     \multicolumn{5}{l}{\textit{Pretrained on large-scale video-text dataset}}  \\  
     \multicolumn{1}{l}{ClipBERT \cite{lei2021less}} & 20.4 & 48.0 & 60.8 & 6.0 \\
     \multicolumn{1}{l}{Frozen \cite{bain2021frozen}} &  31.0 & 59.8 & 72.4 & 3.0 \\
     \multicolumn{1}{l}{HD-VILA \cite{xue2022advancing}} & 28.8 & 57.4 & 69.1 & 4.0 \\
     \multicolumn{1}{l}{All-in-one \cite{wang2022all}} &  32.7 & 61.4 & 73.5 & 3.0 \\
    \multicolumn{1}{l}{BridgeFormer \cite{ge2022bridging}} &  37.0 & 62.2 & 73.9 & 3.0 \\
     \multicolumn{1}{l}{Clover \cite{huang2022clover}} &  48.6 & 74.3 & 82.2 & 2.0 \\ \hline 
     \multicolumn{5}{l}{\textit{CLIP pretrained}}  \\  
     \multicolumn{1}{l}{Clip4clip \cite{luo2021clip4clip}} & 43.4 & 70.2 & 80.6 & 2.0 \\
     \multicolumn{1}{l}{CAMoE* \cite{cheng2021improving}} &  43.8 & 71.4 & - & - \\
     \multicolumn{1}{l}{CLIP2TV \cite{gao2021clip2tv}} &  45.5 & 69.7 & 80.6 & 2.0 \\
     \multicolumn{1}{l}{DRL-B/16 \cite{wang2022disentangled}} &  49.0 & 76.5 & 84.5 & 2.0 \\ \hline 
     \multicolumn{5}{l}{\textit{Our method}}  \\
     \multicolumn{1}{l}{STAN-self-B/32} &  46.2 & 70.4 & 80.0 & 2.0 \\
     \multicolumn{1}{l}{STAN-conv-B/32} &  46.5 & 71.5 & 80.9 & 2.0 \\
     \multicolumn{1}{l}{STAN-conv-B/32*} &  51.3 & 75.1 & 83.4 & 1.0 \\
     \multicolumn{1}{l}{STAN-conv-B/16} &  49.4 & 74.9 & 84.5 & 1.0 \\
     \multicolumn{1}{l}{STAN-conv-B/16*} & \textbf{54.6} & \textbf{78.4} & \textbf{85.1} & 1.0 \\  \hline 
     
   \end{tabular}  }
   \end{center}
   \vspace{-1.7em}
\end{table}

As shown in Table \ref{msrvtt}, \ref{didemo} and \ref{lsmdc}, \textit{CLIP}-based methods generally achieve superior performance than the video-text pretrained methods, which demonstrates the great potential of transferring image-text pretrained models to the video domain. Among the \textit{CLIP}-based methods, our STAN achieves SOTA performance across all three benchmarks at both \textit{CLIP}-B/32 and \textit{CLIP}-B/16 model scales. Specifically, with comparable model size, STAN outperforms the posterior structure based method, \emph{i.e.,} CLIP4clip \cite{luo2021clip4clip} by 2.9\% at R@1 averaged on the three datasets, which shows obvious advantage of the branch structure. Compared to the other SOTAs, \emph{e.g.,} DRL\cite{wang2022disentangled}, which advances video-text retrieval via improving the cross-modality interaction upon visual-language outputs of \textit{CLIP}, STAN shows a different way to achieve competitive performance, which improves the temporal modeling capability of \textit{CLIP} itself. Therefore, empowering \textit{CLIP} model with stronger video representation learning capability, STAN is potentially compatible with the other SOTAs which present advanced techniques operated upon \textit{CLIP} outputs, e.g., hierarchical video-text interaction \cite{wang2022disentangled,min2022hunyuan_tvr} and hard sample modeling\cite{fang2021clip2video}. We leave them for future work. 
Additionally, we also notice that both the self-attention and 3D convolution instantiated model, \emph{i.e.,} STAN-self and STAN-conv, achieve competitive performance with a slight difference. Specifically, STAN-conv is comparable with STAN-self when transferring to smaller datasets, e.g., MSRVTT (-0.3 at R@1) and DiDeMo (+0.3 at R@1) while STAN-self is better on larger scale dataset, e,g., LSMDC (+0.6 at R@1). The results further suggest that self-attention instantiated STAN would be the better choice when transferring \textit{CLIP} to large-scale downstream datasets, while 3D convolution instantiated STAN would be better for the small ones. In Appendix, we present more results with visualization.
    

\begin{table}[]
   \setlength{\abovecaptionskip}{0.cm}
   \begin{center}
     \caption{Comparison on LSMDC \cite{rohrbach2017movie}. * means extra tricks (\textit{e.g.,} DSL \cite{cheng2021improving} and QB-Norm \cite{bogolin2022cross}) are utilized during inference. } 
     
     \footnotesize
     \label{lsmdc}
     \renewcommand\tabcolsep{9pt}
     \scalebox{0.92}{
     \begin{tabular}{ccccc} 
     \hline
     \multicolumn{1}{c}{Methods} & \multicolumn{1}{c}{R@1 $\uparrow$}  & \multicolumn{1}{c}{R@5 $\uparrow$} & \multicolumn{1}{c}{R@10 $\uparrow$} & \multicolumn{1}{c}{MdR $\downarrow$} \\ \hline 
     \multicolumn{5}{l}{\textit{Pretrained on large-scale video-text dataset}}  \\  
     \multicolumn{1}{l}{Frozen \cite{bain2021frozen}} &  15.0 & 30.8 & 40.3 & 20.0 \\
     \multicolumn{1}{l}{HD-VILA \cite{xue2022advancing}} & 17.4 & 34.1 & 44.1 & 15.0 \\
    \multicolumn{1}{l}{BridgeFormer \cite{ge2022bridging}} &  17.9 & 35.4 & 44.5 & 15.0 \\
     \multicolumn{1}{l}{Clover \cite{huang2022clover}} &  22.7 & 42.0 & 52.6 & 9.0 \\ \hline 
     \multicolumn{5}{l}{\textit{CLIP pretrained}}  \\  
     \multicolumn{1}{l}{Clip4Clip \cite{luo2021clip4clip}} & 21.6 & 41.8 & 49.8 & 8.0 \\
     \multicolumn{1}{l}{CAMoE* \cite{cheng2021improving}} &  25.9 & 46.1 & 53.7 & - \\
     \multicolumn{1}{l}{CCLIP-B/16 \cite{zhao2022centerclip}} &  24.2 & 46.2 & 55.9 & 8.0 \\
     \multicolumn{1}{l}{DRL-B/16 \cite{wang2022disentangled}} &  26.5 & 47.6 & 56.8 & 7.0 \\ \hline 
     \multicolumn{5}{l}{\textit{Our method}}  \\
     \multicolumn{1}{l}{STAN-self-B/32} &  23.7 & 42.7 & 51.8 & 9.0 \\
     \multicolumn{1}{l}{STAN-conv-B/32} &  23.1 & 42.2 & 51.0 & 9.0 \\
     \multicolumn{1}{l}{STAN-self-B/32*} &  26.2 & 46.0 & 53.9 & 9.0 \\
     \multicolumn{1}{l}{STAN-self-B/16} &  27.1 & 49.3 & 58.7 & 6.0 \\
     \multicolumn{1}{l}{STAN-self-B/16*} & \textbf{29.2} & \textbf{49.5} & \textbf{58.8} & 6.0 \\  \hline 
     
   \end{tabular}  
   }
   \end{center}
   \vspace{-1.7em}
\end{table} 

\noindent \textbf{Video Recognition.} 
To evaluate the spatial-temporal modeling capability of STAN, we compare it to other SOTAs on video recognition benchmarks, \emph{i.e.,} Kinetics-400 (K400) and Something-Something-v2 (SSv2). The results are reported in Table \ref{k400} and Table \ref{ssv2} respectively. 
On K400 benchmark, \textit{CLIP}-based methods achieve competitive results with much smaller model size compared to the image-pretrained methods, which shows the superiority of image-text pretraining. For example, our VIT-B/16 based STAN outperforms VIT-Huge based ViViT \cite{arnab2021vivit} and Swin3D-L based Video-swin \cite{liu2021video}, which have more than 15× and 88× GFLOPs compared to our method. Meanwhile,  our method achieves SOTA performance among \textit{CLIP}-based methods, which demonstrates the effective of our method on transferring \textit{CLIP} to the video domain.
As for SSv2 benchmark, we find that, without temporal modeling, bare \textit{CLIP} model \cite{dosovitskiy2020image} achieves only 44.0\% top-1 accuracy which dramatically under-performs ImageNet-21K pretrained Timesformer \cite{bulat2021space}, though it owns pretrained knowledge obtained from a much larger image-text dataset. The result suggest that the domain gap is significant between SSv2 and \textit{CLIP} model, and temporal modeling capability is desired for the action recognition task on SSv2. STAN brings about more than 20\% performance improvement over the \textit{CLIP} baseline and achieves competitive compared to other \textit{CLIP}-based methods, which demonstrates that STAN empowers \textit{CLIP} with strong temporal modeling capability.
\begin{table*}[]
   \setlength{\abovecaptionskip}{0.cm}
   \begin{center}
     \caption{Comparison between our method and the state-of-the-arts on Kinetics-400 validation set \cite{kay2017kinetics}. We report the FLOPs of all views.} 
     
     \footnotesize
     \label{k400}
     \renewcommand\tabcolsep{9pt}
     \scalebox{1.05}{
     \begin{tabular}{ccccccc} 
     \hline
     \multicolumn{1}{c}{Methods} & \multicolumn{1}{c}{Pretrain}  & \multicolumn{1}{c}{Frames}  & \multicolumn{1}{c}{Testing Views} & \multicolumn{1}{c}{GFLOPs} & \multicolumn{1}{c}{Top-1 Accuracy} & \multicolumn{1}{c}{Top-5 Accuracy}\\ \hline 
     \multicolumn{7}{l}{\textit{Large-scale image pretraining}}  \\  
     \multicolumn{1}{l}{TimeSformer-L \cite{bertasius2021space}} &  ImageNet-21 K & 96 & $1 \times 3$ & 7140 & 80.7 & 94.7 \\ 
     \multicolumn{1}{l}{Video-Swin-L (384 $\uparrow$) \cite{liu2021video}} &  ImageNet-21 K & 32 & $10 \times 5$ & 105350 & 84.9 & 96.7 \\ 
     \multicolumn{1}{l}{MViTv2-L (312 $\uparrow$) \cite{li2022mvitv2}} &  ImageNet-21 K & 40 & $5 \times 3$ & 42420 & \textbf{86.1} & \textbf{97.0} \\ 
     \multicolumn{1}{l}{ViViT-H \cite{arnab2021vivit}} &  JFT-300M & 32 & $4 \times 3$ & 17352 & 84.8 & 95.8 \\ 
     \multicolumn{1}{l}{TokenLearner-L/10 \cite{ryoo2021tokenlearner}} &  JFT-300M & - & $4 \times 3$ & 48912 & 85.4 & 96.3 \\ 
     \hline 
     \multicolumn{7}{l}{\textit{Large-scale image-text pretraining}}  \\  
     \multicolumn{1}{l}{\textit{CLIP}-B/16 \cite{dosovitskiy2020image}} & CLIP-400M & 8 & $4 \times 3$ & - & 81.1 & 94.8 \\
     \multicolumn{1}{l}{Action-CLIP-B/16 \cite{wang2021actionclip}} & CLIP-400M & 32 & $10 \times 3$ & 16890 & 83.8 & 96.2 \\  
     \multicolumn{1}{l}{A6 \cite{ju2021prompting}} & CLIP-400M & 16 & $ - $ & - & 76.9 & 93.5 \\  
     \multicolumn{1}{l}{STadapter-CLIP-B/16 \cite{wang2021actionclip}} & CLIP-400M & 8 & $1 \times 3$ & 455 & 82.0 & 95.7 \\
     \multicolumn{1}{l}{STadapter-CLIP-B/16 \cite{wang2021actionclip}} & CLIP-400M & 32 & $1 \times 3$ & 1821 & 82.7 & 96.2  \\ 
     \multicolumn{1}{l}{X-CLIP-B/16 \cite{wang2021actionclip}} & CLIP-400M & 8 & $4 \times 3$ & 1740 & 83.8 & 96.7 \\
     \multicolumn{1}{l}{X-CLIP-B/16 \cite{wang2021actionclip}} & CLIP-400M & 16 & $4 \times 3$ & 3444 & 84.7 & \textbf{96.8}  \\ \hline 
     \multicolumn{7}{l}{\textit{Our method}}  \\
     \multicolumn{1}{l}{STAN-conv-B/16} & CLIP-400M  & 8 & $1 \times 3$ & 714  & 83.1 & 96.0 \\
     \multicolumn{1}{l}{STAN-self-B/16} & CLIP-400M  & 8 & $1 \times 3$ & 593  & 84.2 & 96.5 \\
     \multicolumn{1}{l}{STAN-self-B/16} & CLIP-400M  & 16 & $1 \times 3$ & 1187  & \textbf{84.9} & \textbf{96.8}  \\ \hline 
     
   \end{tabular}  }
   \end{center}
   \vspace{-0.5em}
\end{table*}  

\begin{table*}[]
   \setlength{\abovecaptionskip}{0.cm}
   \setlength{\belowcaptionskip}{-0.3cm}
   \begin{center}
     \caption{Comparison on Something-Something-v2 validation set \cite{goyal2017something}. We report the FLOPs of all views. * means our implementation.} 
     
     \footnotesize
     \label{ssv2}
     \renewcommand\tabcolsep{9pt}
     \scalebox{1.05}{
     \begin{tabular}{ccccccc} 
     \hline
     \multicolumn{1}{c}{Methods} & \multicolumn{1}{c}{Pretrain}  & \multicolumn{1}{c}{Frames}  & \multicolumn{1}{c}{Testing Views} & \multicolumn{1}{c}{GFLOPs} & \multicolumn{1}{c}{Top-1 Accuracy} & \multicolumn{1}{c}{Top-5 Accuracy}\\ \hline 
     \multicolumn{1}{l}{TimeSformer-HR \cite{bertasius2021space}} &  ImageNet-21 K & 16 & $1 \times 3$ & 5109 & 62.5 & - \\ 
     \multicolumn{1}{l}{ViViT-L \cite{arnab2021vivit}} &  K400 & 16 & $4 \times 3$ & 11892 & 65.4 & 89.8 \\ 
     \multicolumn{1}{l}{MViT-B-24 \cite{li2022mvitv2}} &  K600 & 32 & $1 \times 3$ & 708 & 68.7 & 91.5 \\ 
     \multicolumn{1}{l}{Video-Swin-B \cite{liu2021video}} &  K400 & 32 & $1 \times 3$ & 963 & \textbf{69.6} & \textbf{92.7} \\ 
     \multicolumn{1}{l}{\textit{CLIP}-B/16 \cite{dosovitskiy2020image}} & CLIP-400M & 8 & $1 \times 3$ & - & 44.0 & 76.2 \\
     \multicolumn{1}{l}{X-CLIP-B/16* \cite{wang2021actionclip}} & CLIP-400M & 8 & $1 \times 3$ & 435 & 63.1 & 89.0 \\
     \multicolumn{1}{l}{STadapter-CLIP-B/16 \cite{wang2021actionclip}} & CLIP-400M & 8 & $1 \times 3$ & 489 & 67.1 & 91.2 \\
     \multicolumn{1}{l}{STadapter-CLIP-B/16 \cite{wang2021actionclip}} & CLIP-400M & 32 & $1 \times 3$ & 1955 & 69.5 & 92.6  \\ \hline 
     \multicolumn{7}{l}{\textit{Our method}}  \\
     \multicolumn{1}{l}{STAN-conv-B/16} & CLIP-400M  & 8 & $1 \times 3$ & 845  & 65.2 & 90.5 \\
     \multicolumn{1}{l}{STAN-self-B/16} & CLIP-400M  & 8 & $1 \times 3$ & 688  & 67.6 & 91.4 \\
     \multicolumn{1}{l}{STAN-self-B/16} & CLIP-400M  & 16 & $1 \times 3$ & 1376  & 69.5 & \textbf{92.7}  \\ \hline 
     
   \end{tabular}  }
   \end{center}
   \vspace{-1.3em}
\end{table*}  

 \begin{table*}[]
    \setlength{\abovecaptionskip}{0.cm}
    \begin{center}
      \caption{Ablation studies on different datasets. For MSRVTT and DiDemo, we use \textit{CLIP}-B/32 as  backbone and report Recall@1; for K400 and SSv2, we use \textit{CLIP}-B/16 as backbone and report Top1 Accuracy. We adopt temporal self-attention here in our Cross-Frame module. } 
      \label{ablation}
      \renewcommand\tabcolsep{9pt}
      \scalebox{1.00}{
      \begin{tabular}{cccccccc} 
      \hline
      \multicolumn{4}{c|}{Components} & \multicolumn{4}{c}{Results} \\ 
      \multicolumn{1}{c}{Cross-Frame}  &  \multicolumn{1}{c}{Intra-Frame} &  \multicolumn{1}{c}{Branch structure} & \multicolumn{1}{c|}{Multi-level} & \multicolumn{1}{c}{MSR-VTT} & \multicolumn{1}{c}{DiDemo} & \multicolumn{1}{c}{K400} & \multicolumn{1}{c}{SSv2}\\ \hline \hline 
       & & & \multicolumn{1}{c|}{}  & 43.1  & 43.4 & 79.9 & 44\\
       $\checkmark$ & $\checkmark$ & & \multicolumn{1}{c|}{} & 44.9 & 43.5 & 80.5 & 55.9\\ 
      $\checkmark$ & $\checkmark$ & $\checkmark$ & \multicolumn{1}{c|}{} & 44.2 & 43.6 & 80.8 & 58.6\\ 
       & $\checkmark$ & $\checkmark$ & \multicolumn{1}{c|}{$\checkmark$}  & 44.3 & 44.5 & 81.0 & 48.1\\ 
      $\checkmark$ &  & $\checkmark$ & \multicolumn{1}{c|}{$\checkmark$}  & 43.1 & 43.7 & 80.0 & 55.7\\ 
      $\checkmark$ & $\checkmark$ & $\checkmark$ & \multicolumn{1}{c|}{$\checkmark$}  & \textbf{46.9}  & \textbf{46.2}  & \textbf{82.6} & \textbf{65.9} \\ \hline 
      \multicolumn{4}{c|}{+ Testing Techniques (DSL \cite{cheng2021improving} or $1 \times 3$-views)} & 49.7  & 51.4 & 84.2 & 67.6\\ \hline
    \end{tabular}  }
    \end{center}
    \vspace{-1.3em}
  \end{table*}
  
\subsection{Ablation Study}
To verify the contribution of different components in our method, we conduct ablation experiments on both video-text retrieval tasks (\emph{i.e.,} MSR-VTT and DiDemo) and video action recognition tasks (\emph{i.e.,} K400 and SSv2).  First of all, according to the results reported in Table \ref{ablation}, we can conclude that components in STAN are compatible with each other while each of them contributes to the transferring of \textit{CLIP}. Specifically, when we remove the branch structure and multi-level feature learning, and append STAN as a posterior structure upon the \textit{CLIP}, the performance of STAN decreased a lot on all four benchmarks, which demonstrates the superiority of our model structure compared to the posterior structure. Besides, we find that without the Cross-Frame module, STAN still brings about performance improvement over baseline, which suggests that our method is beneficial to image-to-video knowledge transferring for \textit{CLIP} model. With the help of Cross-Frame module, the complete STAN further outperforms the baseline by a larger margin and achieves SOTA performance on both video-text retrieval and video recognition tasks, which reveals our method is able to model temporal dependency while taking advantage of knowledge in different level of representation. 

\subsection{Further discussion on STAN} \label{discussion}
\noindent \textbf{The effect of different temporal modeling structures.} 
As aforementioned, posterior structure based temporal modeling for \textit{CLIP} transferring is popular for high-level knowledge dominant tasks, \emph{e.g.,} video-text retrieval, while intermediate structures are employed for low-level knowledge dominant tasks, \emph{e.g.,} video recognition. In section \ref{sec:sota}, we demonstrate the superiority of our branch structured based method compared to other structures on different tasks, respectively. Here, we further adopt posterior structures for video recognition and intermediate structure for video-text retrieval to better understand the effect of different temporal modeling structures. For posterior structures, we employ the Sequential Transformer (Seq Trans) in CLIP4clip \cite{luo2021clip4clip} and Temporal Differential Block (TDB) in CLIP2video \cite{fang2021clip2video}. For intermediate structures, we choose the Message Token (Msg Token) in XCLIP \cite{ni2022expanding} and ST-adapter \cite{pan2022st}. Note that, for a fair comparison, we only report the performance achieved by the temporal modeling structures without other extra techniques (\textit{e.g.}, prompting modeling in \cite{fang2021clip2video} and \cite{ni2022expanding} ). As shown in Table \ref{differstyle}, posterior structures are more effective than intermediate structures in transferring \textit{CLIP} to video-text retrieval tasks, but brings trivial improvement on the video recognition task. In contrast, intermediate structures perform well on video recognition, but bring little improvement over baseline on video-text retrieval.  As for our branch structure based STAN, it not only successfully extends \textit{CLIP} to both tasks but also outperforms both the other two structure based methods, which demonstrates that our structure is a better temporal modeling method in the context of \textit{CLIP}-based image-to-video knowledge transferring.

\begin{table}[]
\renewcommand\arraystretch{1.2}
   \setlength{\abovecaptionskip}{0.cm}
   \begin{center}
     \caption{Analysis of different structures on both video recognition and retrieval tasks. * means our own implementation.} 
     
     \footnotesize
     \label{differstyle}
     \renewcommand\tabcolsep{9pt}
     \scalebox{1.0}{
     \begin{tabular}{cccc} 
     \hline
     \multicolumn{1}{c|}{Method} & \multicolumn{1}{c|}{Style} & \multicolumn{1}{c}{MSR-VTT} & \multicolumn{1}{c}{K400} \\ \hline 
     \multicolumn{1}{l|}{baseline} & \multicolumn{1}{c|}{-} & \multicolumn{1}{c}{43.1} & \multicolumn{1}{c}{79.9} \\ \hline 
     \multicolumn{1}{l|}{Msg Token \cite{ni2022expanding}} & \multicolumn{1}{c|}{Intermediate} & \multicolumn{1}{c}{43.2*} & \multicolumn{1}{c}{82.7} \\
     \multicolumn{1}{l|}{ST-adapter \cite{pan2022st}} & \multicolumn{1}{c|}{Intermediate} & \multicolumn{1}{c}{42.5*} & \multicolumn{1}{c}{82.0} \\ \hline 
     \multicolumn{1}{l|}{Seq Trans \cite{luo2021clip4clip}} & \multicolumn{1}{c|}{Posterior} & \multicolumn{1}{c}{44.5} & \multicolumn{1}{c}{80.5*} \\
     \multicolumn{1}{l|}{TDB \cite{fang2021clip2video}} & \multicolumn{1}{c|}{Posterior} & \multicolumn{1}{c}{45.1} & \multicolumn{1}{c}{81.1*} \\ \hline 
     \multicolumn{1}{l|}{STAN-selfN} & \multicolumn{1}{c|}{Branch} & \multicolumn{1}{c}{\textbf{46.9}} & \multicolumn{1}{c}{\textbf{84.2}} \\ \hline 
     
   \end{tabular}  }
   \end{center}
   \vspace{-1em}
\end{table}

\begin{table}[]
\renewcommand\arraystretch{1.2}
   \setlength{\abovecaptionskip}{0.cm}
   \begin{center}
     \caption{The impact of different levels of inputs from \textit{CLIP} layers on STAN.} 
     
     \footnotesize
     \label{location}
     \renewcommand\tabcolsep{9pt}
     \scalebox{1.0}{
     \begin{tabular}{cccc} 
     \hline
     \multicolumn{2}{c|}{Methods} & \multicolumn{1}{c|}{Recall@1 on MSR-VTT} & Top1 Acc on SSv2\\ \hline 
     \multicolumn{2}{c|}{Baseline} & \multicolumn{1}{c|}{43.1} & 44.0 \\ \hline 
      \multicolumn{1}{c}{\multirow{3}{*}{\rotatebox{90}{Interval}}} & \multicolumn{1}{c|}{3} & \multicolumn{1}{c|}{43.5} & 54.4 \\
      & \multicolumn{1}{c|}{2} & \multicolumn{1}{c|}{44.2} & 61.1 \\
      & \multicolumn{1}{c|}{1} & \multicolumn{1}{c|}{\textbf{46.9}} & \textbf{65.2} \\ \hline 
      \multicolumn{1}{c}{\multirow{3}{*}{\rotatebox{90}{Range}}} & \multicolumn{1}{c|}{1-4} & \multicolumn{1}{c|}{43.4} & 62.2 \\
      & \multicolumn{1}{c|}{5-8} & \multicolumn{1}{c|}{43.9} & 62.3 \\
      & \multicolumn{1}{c|}{9-12} & \multicolumn{1}{c|}{\textbf{46.9}} & \textbf{65.2} \\ \hline 
      
   \end{tabular}  }
   \end{center}
   \vspace{-2em}
\end{table}  

\vspace{0.3em}

\noindent \textbf{The impact of multi-level inputs from \textit{CLIP} layers.} 
STAN acts as a new branch beside the \textit{CLIP} backbone, which takes the video frame representation at different levels of \textit{CLIP} layers as inputs. To study the impact of the choice of different \textit{CLIP} representations, we fixed the number of STAN layers to 4 and vary the level range and interval of selected \textit{CLIP} layers. 
For the interval, we align the last layer of \textit{CLIP} and STAN, and vary the interval between the selected \textit{CLIP} layers. For example, interval=2 means STAN receives outputs of every 2 \textit{CLIP} layers as inputs, \emph{i.e.,} the 6$th$, 8$th$, 10$th$, and 12$th$ layers. As shown in Table \ref{location}, interval=1 is the best choice for both datasets. Then, we fix the interval to 1, and vary the level range of selected \textit{CLIP} layers. The result suggests that the mid-high level of pretrained \textit{CLIP} representation is more valuable for downstream tasks.

\vspace{0.3em}

\begin{figure}[] 
    \setlength{\abovecaptionskip}{-0.01 cm}
    \centering
    \includegraphics[width=0.5\textwidth]{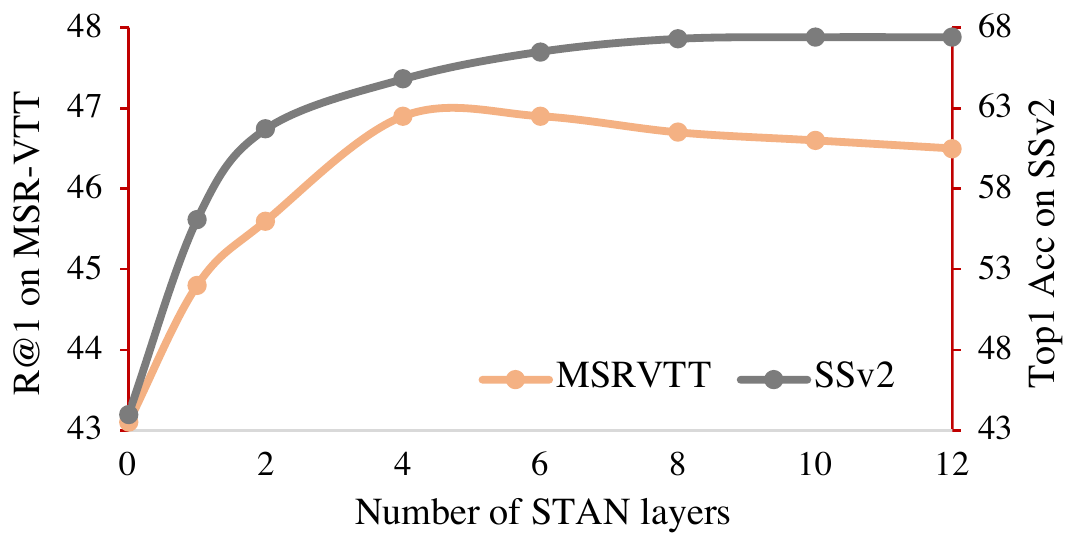} 
    \captionsetup{font={footnotesize}}
    \caption{Analysis of the number of layers in STAN. We report the Recall@1 on MSR-VTT and Top1 Accuracy on SSv2 respectively.}
    \label{num_layers}
    \vspace{-1em}
  \end{figure}
  
\noindent \textbf{The impact of STAN layer number.} 
Intuitively, increasing layers in STAN would access more \textit{CLIP} representation levels bringing about stronger temporal modeling capability. Nevertheless, it also increases the risk of over-fitting to downstream tasks. Thereby, we study the effect of layer number to find a better trade-off. As shown in Fig. \ref{num_layers}, for MSR-VTT retrieval, 
the performance improvement of STAN reaches the peak at 4 layers, and the performance drops with further increasing of layers. On SSv2, the performance improvement of STAN tend to converge after 6 layers. Generally, STAN with 4 to 6 layers is a good choice for different tasks.


\section{Conclusion} 
In this paper, we study the temporal modeling in \textit{CLIP}-based image-to-video knowledge transferring. We first uncover that current methods fail to work on high-level semantic-dominant tasks and low-level visual pattern-dominant tasks simultaneously. Then, to address this problem, we propose a simple and effective framework named Spatial-Temporal Auxiliary Network (STAN) to expand \textit{CLIP} to diverse video tasks. Extensive experiments on Video-Text Retrieval and Video Recognition tasks demonstrate the superiority of our method. 

\vspace{0.5em}
    
\footnotesize{*: equal contribution  
 
  $\Diamond$: Work done when interning at ByteDance Inc}
    
    \clearpage

{\small
\bibliographystyle{ieee_fullname}
\bibliography{egbib}

\begin{thebibliography}{10}\itemsep=-1pt

\bibitem{anne2017localizing}
Lisa Anne~Hendricks, Oliver Wang, Eli Shechtman, Josef Sivic, Trevor Darrell,
  and Bryan Russell.
\newblock Localizing moments in video with natural language.
\newblock In {\em Proceedings of the IEEE international conference on computer
  vision}, pages 5803--5812, 2017.

\bibitem{arnab2021vivit}
Anurag Arnab, Mostafa Dehghani, Georg Heigold, Chen Sun, Mario Lu{\v{c}}i{\'c},
  and Cordelia Schmid.
\newblock Vivit: A video vision transformer.
\newblock In {\em Proceedings of the IEEE/CVF International Conference on
  Computer Vision}, pages 6836--6846, 2021.

\bibitem{bain2021frozen}
Max Bain, Arsha Nagrani, G{\"u}l Varol, and Andrew Zisserman.
\newblock Frozen in time: A joint video and image encoder for end-to-end
  retrieval.
\newblock In {\em Proceedings of the IEEE/CVF International Conference on
  Computer Vision}, pages 1728--1738, 2021.

\bibitem{bertasius2021space}
Gedas Bertasius, Heng Wang, and Lorenzo Torresani.
\newblock Is space-time attention all you need for video understanding?
\newblock In {\em ICML}, volume~2, page~4, 2021.

\bibitem{bogolin2022cross}
Simion-Vlad Bogolin, Ioana Croitoru, Hailin Jin, Yang Liu, and Samuel Albanie.
\newblock Cross modal retrieval with querybank normalisation.
\newblock In {\em Proceedings of the IEEE/CVF Conference on Computer Vision and
  Pattern Recognition}, pages 5194--5205, 2022.

\bibitem{bulat2021space}
Adrian Bulat, Juan~Manuel Perez~Rua, Swathikiran Sudhakaran, Brais Martinez,
  and Georgios Tzimiropoulos.
\newblock Space-time mixing attention for video transformer.
\newblock {\em Advances in Neural Information Processing Systems},
  34:19594--19607, 2021.

\bibitem{carreira2017quo}
Joao Carreira and Andrew Zisserman.
\newblock Quo vadis, action recognition? a new model and the kinetics dataset.
\newblock In {\em proceedings of the IEEE Conference on Computer Vision and
  Pattern Recognition}, pages 6299--6308, 2017.

\bibitem{chen2011collecting}
David Chen and William~B Dolan.
\newblock Collecting highly parallel data for paraphrase evaluation.
\newblock In {\em Proceedings of the 49th annual meeting of the association for
  computational linguistics: human language technologies}, pages 190--200,
  2011.

\bibitem{cheng2021improving}
Xing Cheng, Hezheng Lin, Xiangyu Wu, Fan Yang, and Dong Shen.
\newblock Improving video-text retrieval by multi-stream corpus alignment and
  dual softmax loss.
\newblock {\em arXiv preprint arXiv:2109.04290}, 2021.

\bibitem{dosovitskiy2020image}
Alexey Dosovitskiy, Lucas Beyer, Alexander Kolesnikov, Dirk Weissenborn,
  Xiaohua Zhai, Thomas Unterthiner, Mostafa Dehghani, Matthias Minderer, Georg
  Heigold, Sylvain Gelly, et~al.
\newblock An image is worth 16x16 words: Transformers for image recognition at
  scale.
\newblock {\em arXiv preprint arXiv:2010.11929}, 2020.

\bibitem{fan2021can}
Quanfu Fan, Chun-Fu Chen, and Rameswar Panda.
\newblock Can an image classifier suffice for action recognition?
\newblock In {\em International Conference on Learning Representations}, 2021.

\bibitem{fang2021clip2video}
Han Fang, Pengfei Xiong, Luhui Xu, and Yu Chen.
\newblock Clip2video: Mastering video-text retrieval via image clip.
\newblock {\em arXiv preprint arXiv:2106.11097}, 2021.

\bibitem{gao2021clip2tv}
Zijian Gao, Jingyu Liu, Sheng Chen, Dedan Chang, Hao Zhang, and Jinwei Yuan.
\newblock Clip2tv: An empirical study on transformer-based methods for
  video-text retrieval.
\newblock {\em arXiv preprint arXiv:2111.05610}, 2021.

\bibitem{ge2022bridging}
Yuying Ge, Yixiao Ge, Xihui Liu, Dian Li, Ying Shan, Xiaohu Qie, and Ping Luo.
\newblock Bridging video-text retrieval with multiple choice questions.
\newblock In {\em Proceedings of the IEEE/CVF Conference on Computer Vision and
  Pattern Recognition}, pages 16167--16176, 2022.

\bibitem{gorti2022x}
Satya~Krishna Gorti, No{\"e}l Vouitsis, Junwei Ma, Keyvan Golestan, Maksims
  Volkovs, Animesh Garg, and Guangwei Yu.
\newblock X-pool: Cross-modal language-video attention for text-video
  retrieval.
\newblock In {\em Proceedings of the IEEE/CVF Conference on Computer Vision and
  Pattern Recognition}, pages 5006--5015, 2022.

\bibitem{goyal2017something}
Raghav Goyal, Samira Ebrahimi~Kahou, Vincent Michalski, Joanna Materzynska,
  Susanne Westphal, Heuna Kim, Valentin Haenel, Ingo Fruend, Peter Yianilos,
  Moritz Mueller-Freitag, et~al.
\newblock The" something something" video database for learning and evaluating
  visual common sense.
\newblock In {\em Proceedings of the IEEE international conference on computer
  vision}, pages 5842--5850, 2017.

\bibitem{huang2022clover}
Jingjia Huang, Yinan Li, Jiashi Feng, Xiaoshuai Sun, and Rongrong Ji.
\newblock Clover: Towards a unified video-language alignment and fusion model.
\newblock {\em arXiv preprint arXiv:2207.07885}, 2022.

\bibitem{jia2021scaling}
Chao Jia, Yinfei Yang, Ye Xia, Yi-Ting Chen, Zarana Parekh, Hieu Pham, Quoc Le,
  Yun-Hsuan Sung, Zhen Li, and Tom Duerig.
\newblock Scaling up visual and vision-language representation learning with
  noisy text supervision.
\newblock In {\em International Conference on Machine Learning}, pages
  4904--4916. PMLR, 2021.

\bibitem{ju2021prompting}
Chen Ju, Tengda Han, Kunhao Zheng, Ya Zhang, and Weidi Xie.
\newblock Prompting visual-language models for efficient video understanding.
\newblock {\em arXiv preprint arXiv:2112.04478}, 2021.

\bibitem{kay2017kinetics}
Will Kay, Joao Carreira, Karen Simonyan, Brian Zhang, Chloe Hillier, Sudheendra
  Vijayanarasimhan, Fabio Viola, Tim Green, Trevor Back, Paul Natsev, et~al.
\newblock The kinetics human action video dataset.
\newblock {\em arXiv preprint arXiv:1705.06950}, 2017.

\bibitem{lei2021less}
Jie Lei, Linjie Li, Luowei Zhou, Zhe Gan, Tamara~L Berg, Mohit Bansal, and
  Jingjing Liu.
\newblock Less is more: Clipbert for video-and-language learning via sparse
  sampling.
\newblock In {\em Proceedings of the IEEE/CVF Conference on Computer Vision and
  Pattern Recognition}, pages 7331--7341, 2021.

\bibitem{li2022mvitv2}
Yanghao Li, Chao-Yuan Wu, Haoqi Fan, Karttikeya Mangalam, Bo Xiong, Jitendra
  Malik, and Christoph Feichtenhofer.
\newblock Mvitv2: Improved multiscale vision transformers for classification
  and detection.
\newblock In {\em Proceedings of the IEEE/CVF Conference on Computer Vision and
  Pattern Recognition}, pages 4804--4814, 2022.

\bibitem{Lin_2017_CVPR}
Tsung-Yi Lin, Piotr Dollar, Ross Girshick, Kaiming He, Bharath Hariharan, and
  Serge Belongie.
\newblock Feature pyramid networks for object detection.
\newblock In {\em Proceedings of the IEEE Conference on Computer Vision and
  Pattern Recognition (CVPR)}, July 2017.

\bibitem{liu2022contextual}
Ruyang Liu, Hao Liu, Ge Li, Haodi Hou, TingHao Yu, and Tao Yang.
\newblock Contextual debiasing for visual recognition with causal mechanisms.
\newblock In {\em Proceedings of the IEEE/CVF Conference on Computer Vision and
  Pattern Recognition}, pages 12755--12765, 2022.

\bibitem{liu2021video}
Ze Liu, Jia Ning, Yue Cao, Yixuan Wei, Zheng Zhang, Stephen Lin, and Han Hu.
\newblock Video swin transformer.
\newblock {\em arXiv preprint arXiv:2106.13230}, 2021.

\bibitem{luo2021clip4clip}
Huaishao Luo, Lei Ji, Ming Zhong, Yang Chen, Wen Lei, Nan Duan, and Tianrui Li.
\newblock Clip4clip: An empirical study of clip for end to end video clip
  retrieval.
\newblock {\em arXiv preprint arXiv:2104.08860}, 2021.

\bibitem{miech2019howto100m}
Antoine Miech, Dimitri Zhukov, Jean-Baptiste Alayrac, Makarand Tapaswi, Ivan
  Laptev, and Josef Sivic.
\newblock Howto100m: Learning a text-video embedding by watching hundred
  million narrated video clips.
\newblock In {\em Proceedings of the IEEE/CVF International Conference on
  Computer Vision}, pages 2630--2640, 2019.

\bibitem{min2022hunyuan_tvr}
Shaobo Min, Weijie Kong, Rong-Cheng Tu, Dihong Gong, Chengfei Cai, Wenzhe Zhao,
  Chenyang Liu, Sixiao Zheng, Hongfa Wang, Zhifeng Li, et~al.
\newblock Hunyuan\_tvr for text-video retrivial.
\newblock {\em arXiv preprint arXiv:2204.03382}, 2022.

\bibitem{ni2022expanding}
Bolin Ni, Houwen Peng, Minghao Chen, Songyang Zhang, Gaofeng Meng, Jianlong Fu,
  Shiming Xiang, and Haibin Ling.
\newblock Expanding language-image pretrained models for general video
  recognition.
\newblock {\em arXiv preprint arXiv:2208.02816}, 2022.

\bibitem{pan2022st}
Junting Pan, Ziyi Lin, Xiatian Zhu, Jing Shao, and Hongsheng Li.
\newblock St-adapter: Parameter-efficient image-to-video transfer learning for
  action recognition.
\newblock {\em arXiv preprint arXiv:2206.13559}, 2022.

\bibitem{patashnik2021styleclip}
Or Patashnik, Zongze Wu, Eli Shechtman, Daniel Cohen-Or, and Dani Lischinski.
\newblock Styleclip: Text-driven manipulation of stylegan imagery.
\newblock In {\em Proceedings of the IEEE/CVF International Conference on
  Computer Vision}, pages 2085--2094, 2021.

\bibitem{radford2021learning}
Alec Radford, Jong~Wook Kim, Chris Hallacy, Aditya Ramesh, Gabriel Goh,
  Sandhini Agarwal, Girish Sastry, Amanda Askell, Pamela Mishkin, Jack Clark,
  et~al.
\newblock Learning transferable visual models from natural language
  supervision.
\newblock In {\em International Conference on Machine Learning}, pages
  8748--8763. PMLR, 2021.

\bibitem{rohrbach2017movie}
Anna Rohrbach, Atousa Torabi, Marcus Rohrbach, Niket Tandon, Christopher Pal,
  Hugo Larochelle, Aaron Courville, and Bernt Schiele.
\newblock Movie description.
\newblock {\em International Journal of Computer Vision}, 123(1):94--120, 2017.

\bibitem{ryoo2021tokenlearner}
Michael Ryoo, AJ Piergiovanni, Anurag Arnab, Mostafa Dehghani, and Anelia
  Angelova.
\newblock Tokenlearner: Adaptive space-time tokenization for videos.
\newblock {\em Advances in Neural Information Processing Systems},
  34:12786--12797, 2021.

\bibitem{Sevilla-Lara_2021_WACV}
Laura Sevilla-Lara, Shengxin Zha, Zhicheng Yan, Vedanuj Goswami, Matt Feiszli,
  and Lorenzo Torresani.
\newblock Only time can tell: Discovering temporal data for temporal modeling.
\newblock In {\em Proceedings of the IEEE/CVF Winter Conference on Applications
  of Computer Vision (WACV)}, pages 535--544, January 2021.

\bibitem{sun2019learning}
Chen Sun, Fabien Baradel, Kevin Murphy, and Cordelia Schmid.
\newblock Learning video representations using contrastive bidirectional
  transformer.
\newblock {\em arXiv preprint arXiv:1906.05743}, 2019.

\bibitem{sun2019videobert}
Chen Sun, Austin Myers, Carl Vondrick, Kevin Murphy, and Cordelia Schmid.
\newblock Videobert: A joint model for video and language representation
  learning.
\newblock In {\em Proceedings of the IEEE/CVF International Conference on
  Computer Vision}, pages 7464--7473, 2019.

\bibitem{tran2015learning}
Du Tran, Lubomir Bourdev, Rob Fergus, Lorenzo Torresani, and Manohar Paluri.
\newblock Learning spatiotemporal features with 3d convolutional networks.
\newblock In {\em Proceedings of the IEEE international conference on computer
  vision}, pages 4489--4497, 2015.

\bibitem{vaswani2017attention}
Ashish Vaswani, Noam Shazeer, Niki Parmar, Jakob Uszkoreit, Llion Jones,
  Aidan~N Gomez, {\L}ukasz Kaiser, and Illia Polosukhin.
\newblock Attention is all you need.
\newblock {\em Advances in neural information processing systems}, 30, 2017.

\bibitem{wang2022all}
Alex~Jinpeng Wang, Yixiao Ge, Rui Yan, Yuying Ge, Xudong Lin, Guanyu Cai,
  Jianping Wu, Ying Shan, Xiaohu Qie, and Mike~Zheng Shou.
\newblock All in one: Exploring unified video-language pre-training.
\newblock {\em arXiv preprint arXiv:2203.07303}, 2022.

\bibitem{wang2021actionclip}
Mengmeng Wang, Jiazheng Xing, and Yong Liu.
\newblock Actionclip: A new paradigm for video action recognition.
\newblock {\em arXiv preprint arXiv:2109.08472}, 2021.

\bibitem{wang2022disentangled}
Qiang Wang, Yanhao Zhang, Yun Zheng, Pan Pan, and Xian-Sheng Hua.
\newblock Disentangled representation learning for text-video retrieval.
\newblock {\em arXiv preprint arXiv:2203.07111}, 2022.

\bibitem{xie2018rethinking}
Saining Xie, Chen Sun, Jonathan Huang, Zhuowen Tu, and Kevin Murphy.
\newblock Rethinking spatiotemporal feature learning: Speed-accuracy trade-offs
  in video classification.
\newblock In {\em Proceedings of the European conference on computer vision
  (ECCV)}, pages 305--321, 2018.

\bibitem{xu2016msr}
Jun Xu, Tao Mei, Ting Yao, and Yong Rui.
\newblock Msr-vtt: A large video description dataset for bridging video and
  language.
\newblock In {\em Proceedings of the IEEE conference on computer vision and
  pattern recognition}, pages 5288--5296, 2016.

\bibitem{xue2022advancing}
Hongwei Xue, Tiankai Hang, Yanhong Zeng, Yuchong Sun, Bei Liu, Huan Yang,
  Jianlong Fu, and Baining Guo.
\newblock Advancing high-resolution video-language representation with
  large-scale video transcriptions.
\newblock In {\em Proceedings of the IEEE/CVF Conference on Computer Vision and
  Pattern Recognition}, pages 5036--5045, 2022.

\bibitem{yuan2021florence}
Lu Yuan, Dongdong Chen, Yi-Ling Chen, Noel Codella, Xiyang Dai, Jianfeng Gao,
  Houdong Hu, Xuedong Huang, Boxin Li, Chunyuan Li, et~al.
\newblock Florence: A new foundation model for computer vision.
\newblock {\em arXiv preprint arXiv:2111.11432}, 2021.

\bibitem{yuan2021tokens}
Li Yuan, Yunpeng Chen, Tao Wang, Weihao Yu, Yujun Shi, Zi-Hang Jiang,
  Francis~EH Tay, Jiashi Feng, and Shuicheng Yan.
\newblock Tokens-to-token vit: Training vision transformers from scratch on
  imagenet.
\newblock In {\em Proceedings of the IEEE/CVF International Conference on
  Computer Vision}, pages 558--567, 2021.

\bibitem{zhao2022centerclip}
Shuai Zhao, Linchao Zhu, Xiaohan Wang, and Yi Yang.
\newblock Centerclip: Token clustering for efficient text-video retrieval.
\newblock {\em arXiv preprint arXiv:2205.00823}, 2022.

\bibitem{zhou2022learning}
Kaiyang Zhou, Jingkang Yang, Chen~Change Loy, and Ziwei Liu.
\newblock Learning to prompt for vision-language models.
\newblock {\em International Journal of Computer Vision}, 130(9):2337--2348,
  2022.

\end{thebibliography}
}

\clearpage
\appendix

{\normalsize
\begin{table}[]
   \setlength{\abovecaptionskip}{0.cm}
   \begin{center}
    \caption{Comparisons on MSVD \cite{chen2011collecting}.  * means DSL \cite{cheng2021improving}  is utilized during inference. We don't implement DSL \cite{cheng2021improving} on MSVD due to multiple captions.}  
     \footnotesize
     \label{msvd}
     \renewcommand\tabcolsep{9pt}
     \scalebox{0.90}{
     \begin{tabular}{ccccc} 
     \hline
     \multicolumn{1}{c}{Methods} & \multicolumn{1}{c}{R@1 $\uparrow$}  & \multicolumn{1}{c}{R@5 $\uparrow$} & \multicolumn{1}{c}{R@10 $\uparrow$} & \multicolumn{1}{c}{MdR $\downarrow$} \\ \hline 
     \multicolumn{5}{l}{\textit{Pretrained on large-scale video-text dataset}}  \\  
     \multicolumn{1}{l}{Frozen \cite{bain2021frozen}} & 45.6 & 79.8 & 88.2 & 2 \\
    \multicolumn{1}{l}{BridgeFormer \cite{ge2022bridging}} &  52.0 & 82.8 & 90.0 & 2.0 \\  \hline 
     \multicolumn{5}{l}{\textit{CLIP pretrained}}  \\  
     \multicolumn{1}{l}{Clip4Clip \cite{luo2021clip4clip}} & 46.2 & 76.1 & 84.6 & 2.0 \\
     \multicolumn{1}{l}{CAMoE* \cite{cheng2021improving}} &  49.8 & 79.2 & 87.0 & - \\
     \multicolumn{1}{l}{DRL-B/16 \cite{wang2022disentangled}} &  50.0 & 81.5 & 89.5 & 2.0 \\
     \multicolumn{1}{l}{CenterCLIP-B/16 \cite{zhao2022centerclip}} &  50.6 & 80.3 & 88.4 & 1.0 \\ \hline 
     \multicolumn{5}{l}{\textit{Our method}}  \\
     \multicolumn{1}{l}{STAN-self-B/32} &  46.7 & 76.1 & 86.0 & 2.0 \\
     \multicolumn{1}{l}{STAN-conv-B/32} &  47.5 & 77.6 & 86.5 & 2.0 \\
     \multicolumn{1}{l}{STAN-conv-B/16} &  51.5 & 80.4 & 88.5 & 1.0 \\  \hline 
   \end{tabular}  }
   \end{center}
\end{table}

\section{More supplementary experiments }
\subsection{Comparisons on MSVD}
\textit{MSVD} is a small scale video-text retrieval benchmark, which contains only 1970 videos in total. Each video has averaged 40 corresponding captions. All videos are split into 1200, 100, and 670 for training, validation, and testing. Compared with other retrieval datasets, MSVD contains much fewer training samples with much more captions for each video.  

 As reported in Table \ref{msvd}, CLIP-based methods are less competitive than video pretraining. The results may arise from the much fewer training videos, which makes it harder for \textit{CLIP}-based methods to learn basic video patterns. Our method achieves best performance among CLIP-based methods. Besides, on MSVD, the convolution-based temporal modeling based STAN-conv is superior to the self-attention based STAN-self , which further verifies that STAN-conv could be a better choice when transferring CLIP to small scale downstream datasets.

\vspace{0.5em}

 \subsection{Performance of STAN with different instantiations of temporal modeling }
Our STAN is potentially flexible to different implementations of the cross-frame module for temporal modeling. In our paper, we investigate the performance of STAN with the two most popular implementations, i.e., convolution based and self-attention-based modules. Here, we further present STAN with another two common temporal modeling strategies \emph{i.e.,} 3-D window attention \cite{liu2021video} and message token \cite{ni2022expanding}. As shown in Table \ref{DTM}, all different implementations of temporal modeling modules brings about performance improvement over the CLIP4clip, while the self-attention based implementation achieves the best performance across the three benchmarks.

\begin{table}[]
\renewcommand\arraystretch{1.2}
   \setlength{\abovecaptionskip}{0.cm}
   \begin{center}
     \caption{Different temporal modeling in the cross-frame module. We evaluate four structures on MSR-VTT, DiDemo and K400.} 
     
     \footnotesize
     \label{DTM}
     \renewcommand\tabcolsep{9pt}
     \scalebox{0.95}{
     \begin{tabular}{cccc} 
     \hline
     \multicolumn{1}{c|}{Temporal Module} & \multicolumn{1}{c}{MSR-VTT} & \multicolumn{1}{c}{DiDemo} & \multicolumn{1}{c}{SSv2} \\ \hline 
     \multicolumn{1}{l|}{baseline (CLIP4clip)} & 44.5 & 43.4 & 54.7 \\ \hline 
     \multicolumn{1}{l|}{STAN-Self-Attention} & 46.9  & 46.2  & 67.6 \\
     \multicolumn{1}{l|}{STAN-3-D Convolution} & 46.6  & 46.5  & 65.2 \\
     \multicolumn{1}{l|}{STAN-Window Attention \cite{liu2021video}} & 45.3  & 45.5  & 60.2 \\
     \multicolumn{1}{l|}{STAN-Message Token \cite{ni2022expanding}} & 45.9  & 45.0  & 63.3 \\ \hline 
   \end{tabular}  }
   \end{center}
   \vspace{3em}
\end{table}  

\section{Qualitative evaluation on ``knowledge'' and ``temporal modeling'' in CLIP Transferring}
In this paper, we reveal the key factor for successfully extending image-text pretrained models to the video domain, \emph{i.e.,} taking full advantage of pretrained knowledge while empowering the model with temporal modeling capability.
In experiments, we quantitatively demonstrated the superiority of STAN on both high-level knowledge-dominant video-text retrieval tasks and low-level knowledge-dominant video recognition tasks. Here, we further provide qualitative results to reveal the efficacy of STAN on handling the three key factors.

\begin{table*}[t]
\renewcommand\arraystretch{1.2}
   \setlength{\abovecaptionskip}{0.cm}
   \begin{center}
     \caption{Affect of different structures on video-text representation alignments. The experiments are conducted on MSRVTT-1kA test set.} 
     
     \footnotesize
     \label{tab:avg cos_sim}
     \renewcommand\tabcolsep{9pt}
     \scalebox{0.96}{
     \begin{tabular}{cc|cc} 
     \hline
     \multicolumn{1}{c|}{\multirow{2}{*}{Method}} & \multicolumn{1}{c|}{\multirow{2}{*}{Structure}} & \multicolumn{1}{c|}{Averaged \textbf{similarity scores}} & \multicolumn{1}{c}{Averaged \textbf{similarity margin} between}  \\ 
     \multicolumn{1}{c|}{} &\multicolumn{1}{c|}{} &  \multicolumn{1}{c|}{of positive pairs} & positive and negative samples \\ \hline 
     \multicolumn{1}{l|}{Baseline} & - & \multicolumn{1}{c|}{0.44}  & 0.31  \\ \hline 
     \multicolumn{1}{l|}{CLIP4clip-seqTrans} & Posterior& \multicolumn{1}{c|}{0.48}  & 0.31   \\
     \multicolumn{1}{l|}{XCLP} & Intermediate & \multicolumn{1}{c|}{0.38}  & 0.27  \\
     \multicolumn{1}{l|}{STAN}& Branch & \multicolumn{1}{c|}{\textbf{0.50}}  & \textbf{0.34}   \\ \hline 
   \end{tabular}  }
   \end{center}
   \vspace{-0.6em}
\end{table*}  

\begin{figure*}[h] 
    \setlength{\abovecaptionskip}{-0.005 cm}
    \centering
    \includegraphics[width=1\textwidth]{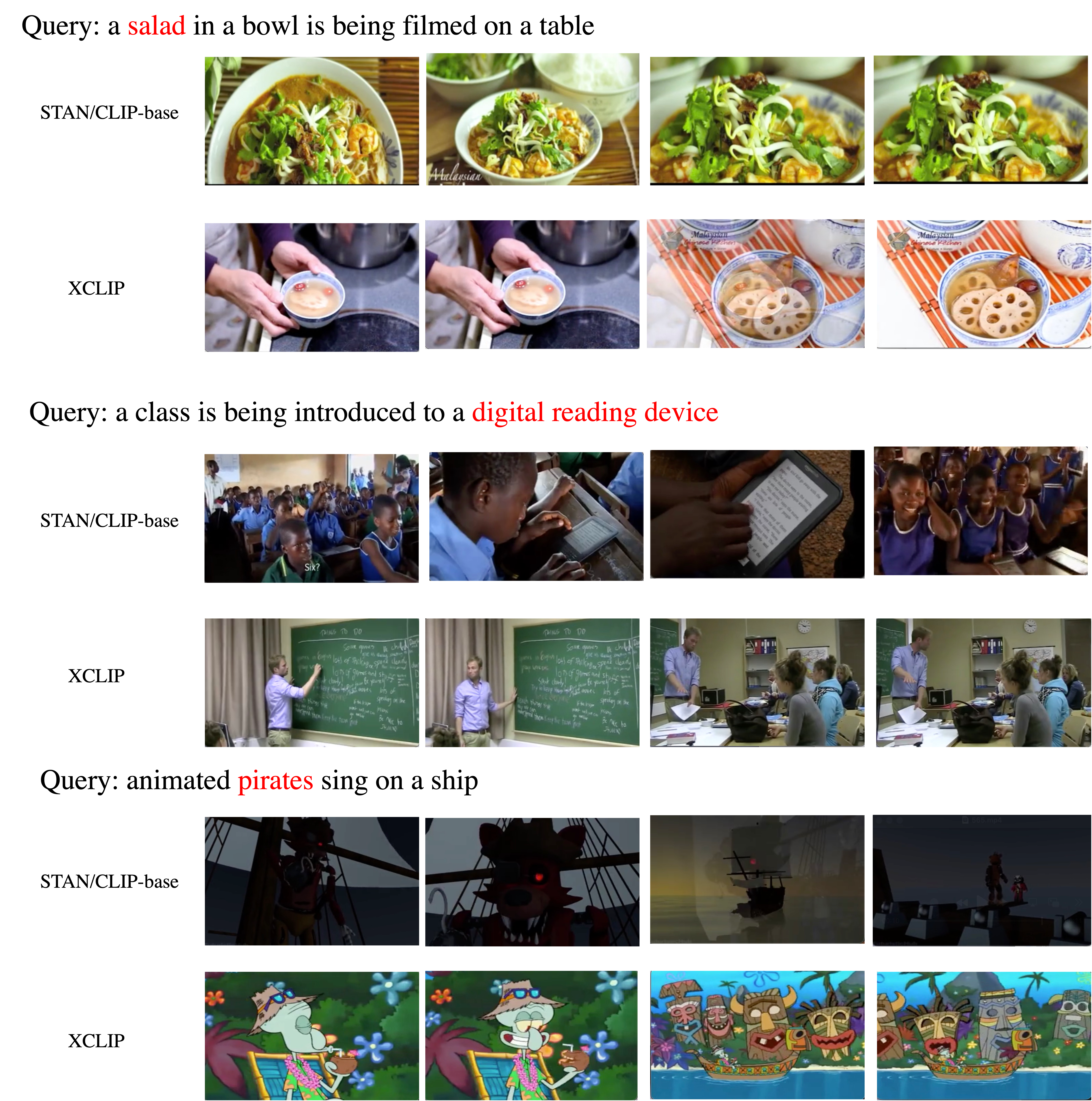} 
    \caption{Qualitative results of text to video retrieval on MSR-VTT. Given a text query, we present the correct matched video returned by STAN and CLIP-base in the first row, and show the false result of XCLIP in the second row. The word highlighted in red indicates the key content missed in the false result.}
    \label{high}
    \vspace{-0.5em}
  \end{figure*}
  
  \begin{figure*}h] 
    \setlength{\abovecaptionskip}{-0.01 cm}
    \centering
    \includegraphics[width=1\textwidth]{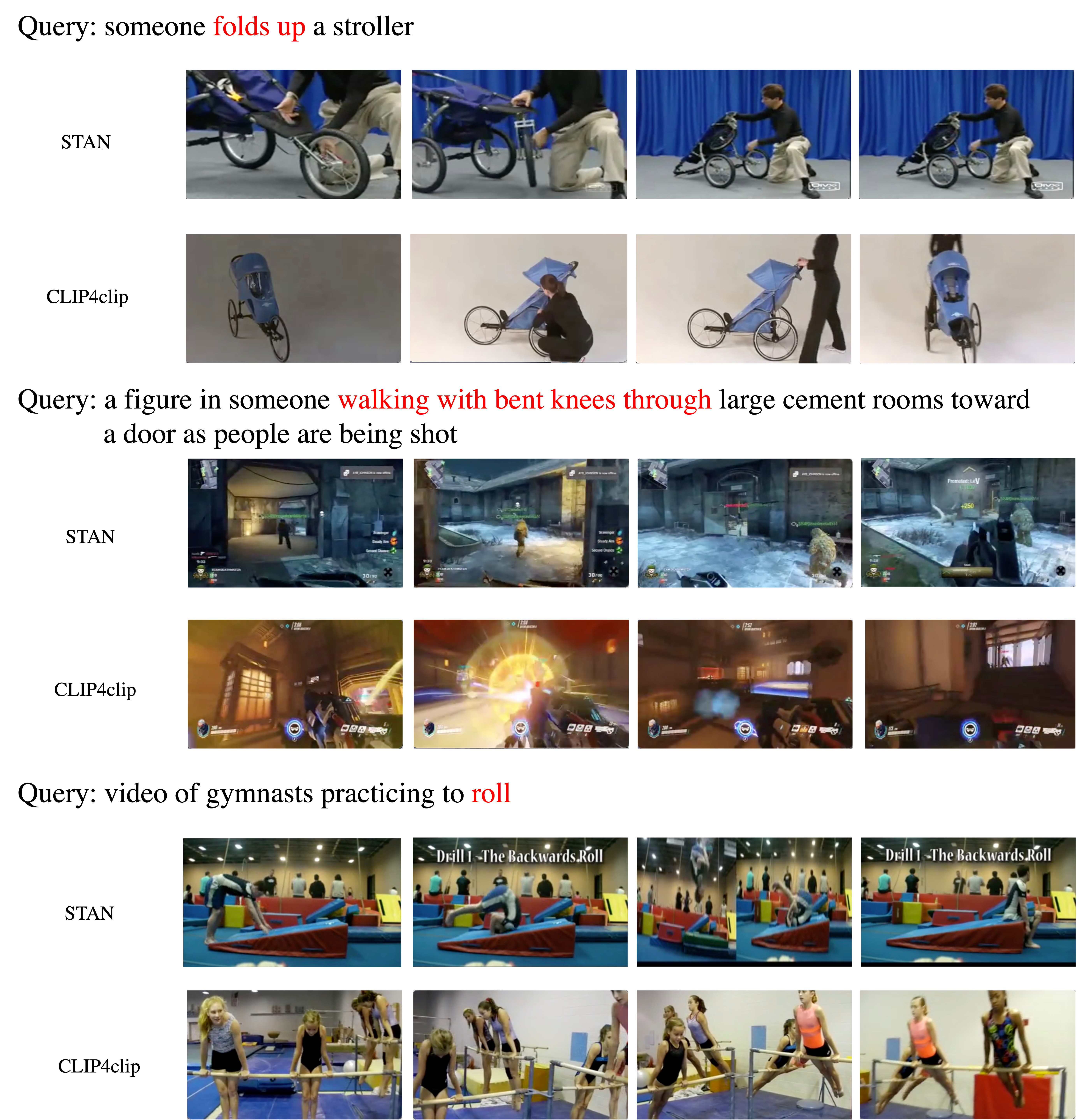} 
    \caption{Qualitative results of text to video retrieval on MSR-VTT. Given a text query, we present the correct matched video returned by STAN in the first row, and show the false result of CLIP4clip in the second row. The word highlighted in red indicates the key content missed in the false result.}
    \label{low}
    \vspace{0.5em}
  \end{figure*}

  \begin{figure*}[h] 
    \setlength{\abovecaptionskip}{-0.01 cm}
    \centering
    \includegraphics[width=1\textwidth]{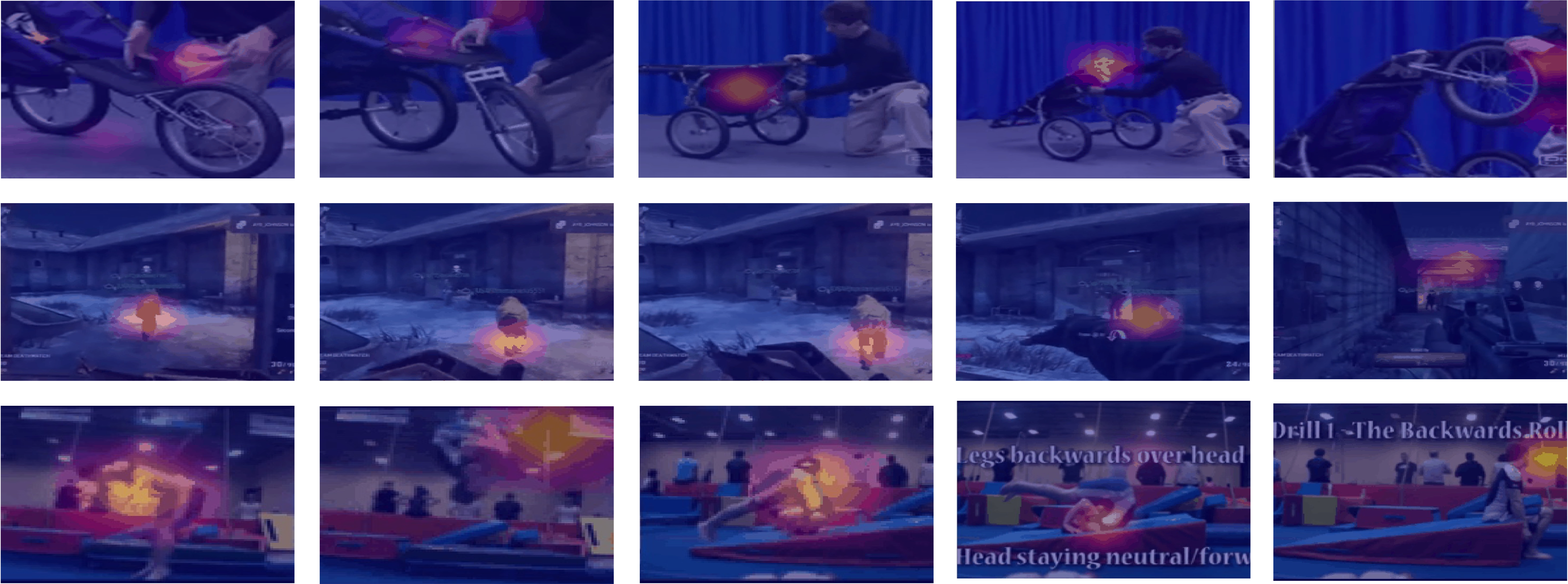} 
    \caption{Visualziation of intra-frame module of STAN on MSR-VTT. Given a text query. The region in red gains more attention from the model. We visualize the attention with VideoCAM.}
   \label{att}
    \vspace{-0.5em}
  \end{figure*}
  
First of all, we present some text-to-video retrieval results of baseline method CLIP-base, intermediate-structure based method XCLIP \cite{ni2022expanding} and  our STAN. 
As shown in the Fig.\ref{high}, these cases could be easily solved if a model can align the highlighted object concepts in queries, \emph{e.g., salad,} with videos containing the visual contents. However, XCLIP returns false results where the key objects are absent from the videos. The results reveal that the intermediate structure can not transfer the high-level visual-text alignment knowledge as well as our method. To further validate the high-level knowledge transferring capability of different methods,  we further report averaged cosine similarity of videos and texts in the MSRVTT dataset calculated by models with different temporal structures. For a fair comparison, we select all video queries (113 in total) that were successfully responded to by all the models. We recall 100 results for each query, and take the ground-truth pairs as positive and others as negative. As shown in  Tab.~\ref{tab:avg cos_sim}, the similarity scores of the XCLIP representation for positive pairs are even lower than the scores of representation from the baseline method. Besides, the similarity margin between the positive samples and the negative samples calculated with XCLIP representation is still the lowest among all the methods. The results reveal that intermediate structure is moderate in learning visual-text aligned representation when transferring CLIP to downstream tasks. In contrast, among all these methods, our method assigns the highest similarity scores to positive pairs and holds the largest margin between the positive and negative. It reveals that with the assistance of STAN, the cross-modal representation is better aligned, where the distance between video and text that contain consistent semantics is  reduced, while the margin between the misaligned samples is increased. 
  
Then, we present some failure text-to-video retrieval cases for CLIP4clip \cite{luo2021clip4clip} in Fig.\ref{low}. As shown in the figure, the posterior structure-based based CLIP4clip returns false results which contain correct static contexts described in queries (\emph{e.g., stroller, gymnasts}) but wrong dynamic information not consistent with the highlighted concepts in queries (\emph{e.g., folds up, roll}). The results show that our method can better leverage the spatial-temporal information for video understanding.

At last, we visualize the attention of the STAN intra-frame module via VideoCAM in Fig.\ref{att}. The results show that our STAN module focuses on the key contents in videos across time.
}
\end{document}